\documentclass[
    journal,
    twocolumn,
    ]{IEEEtranTCOM}
\usepackage{graphicx}
\usepackage{amsmath}
\usepackage{amssymb}
\usepackage{epsfig}
\usepackage{epstopdf}
\usepackage{amsthm}
\usepackage{afterpage}
\usepackage{amsmath,amssymb,amsfonts}
\usepackage{verbatim}
\usepackage{psfrag}
\usepackage{color}
\usepackage[table]{xcolor}
\usepackage{setspace}
\usepackage{adjustbox}
\usepackage{epsfig}
\usepackage{epstopdf}
\usepackage{footmisc}
\usepackage{fmtcount}
\usepackage{stackrel}
\usepackage{float}
\usepackage{breqn}
\usepackage{hyperref}
\usepackage{enumerate}
\usepackage{graphicx}
\usepackage{subcaption}
\usepackage{flushend}
\usepackage{soul}
\usepackage{graphicx}
\usepackage{tikz}
\usetikzlibrary{arrows.meta, positioning, shapes, fit}

\usepackage{tabularx,booktabs}

\begin{document}

\title{Redefining Elderly Care with Agentic AI: Challenges and Opportunities}
\author{Ruhul Amin Khalil, \IEEEmembership{Member, IEEE}; 
Kashif Ahmad, \IEEEmembership{Senior Member, IEEE}; and 
Hazrat Ali, \IEEEmembership{Senior Member, IEEE}
\thanks{Ruhul Amin Khalil is with the Engineering Requirement Unit, College of Engineering, United Arab Emirates University (UAEU), Al-Ain 15551, United Arab Emirates (e-mail: ruhulamin@uaeu.ac.ae).}
\thanks{Kashif Ahmad is with the Department of Computer Science, Munster Technological University, Cork, Ireland (email: kashif.ahmad@mtu.ie)}
\thanks{Hazrat Ali is with the Division of Computing Science and Mathematics, University of Stirling, Stirling, FK9 4LA, UK (e-mail: hazrat.ali@live.com).}
\thanks{Manuscript received \today; revised ~X, X~X.}}

\maketitle

\begin{abstract}
The global ageing population necessitates new and emerging strategies for caring for older adults. In this article, we explore the potential for transformation in elderly care through Agentic Artificial Intelligence (AI), powered by Large Language Models (LLMs). We discuss the proactive and autonomous decision-making facilitated by Agentic AI in elderly care. Personalized tracking of health, cognitive care, and environmental management, all aimed at enhancing independence and high-level living for older adults, represents important areas of application. With a potential for significant transformation of elderly care, Agentic AI also raises profound concerns about data privacy and security, decision independence, and access. We share key insights to emphasize the need for ethical safeguards, privacy protections, and transparent decision-making. Our goal in this article is to provide a balanced discussion of both the potential and the challenges associated with Agentic AI, and to provide insights into its responsible use in elderly care, to bring Agentic AI into harmony with the requirements and vulnerabilities specific to the elderly. Finally, we identify the priorities for the academic research communities, to achieve human-centered advancements and integration of Agentic AI in elderly care. To the best of our knowledge, this is no existing study that reviews the role of Agentic AI in elderly care. Hence, we address the literature gap by analyzing the unique capabilities, applications, and limitations of LLM-based Agentic AI in elderly care. We also provide a companion interactive dashboard at https://hazratali.github.io/agenticai/.
\end{abstract}

\begin{IEEEkeywords}  
Agentic AI, Elderly Care, Digital Health Innovations, Medical Artificial Intelligence, Healthcare AI.
\end{IEEEkeywords}

\section{Introduction}\label{sec:introduction}
\subsection{Overview} 
The United Nations projects the global population to reach approximately 10.4 billion by 2086, before starting a slow decline to about 10.3 billion by 2100 \cite{guillemot2024population}. Estimates for the near future show that by 2030, one in six people worldwide will be over the age of 65, with this number expected to double by 2050 \cite{un2022worlpopulation}. 
The global rise in ageing populations presents a significant challenge for healthcare and social care systems, emphasizing the urgent need for innovative solutions in elderly care. 
For instance, according to the Institute for Fiscal Studies\footnote{https://ifs.org.uk/} report \cite{Antonella2025IFS}, the adult social care system in England faces significant challenges that demand innovative solutions. In England, funding allocation of £24.5 billion for adult social care planned for the year 2024-25 accounted for 40\% of the local authority budget. Of this, approximately half of the budget is consumed by the supporting services for adults aged 65+. Demand for care services among working-age adults grew by 18\% between 2014–15 and 2022–23, outpacing population growth by almost a factor of 3. Despite significant older population growth, state-funded care for older people has dropped by 10\% since 2014–15 due to tightening eligibility criteria \cite{Antonella2025IFS}. Moreover, in the United Kingdom, the number of care worker visa applications has decreased significantly, from 18,300 in August 2023 to 2,300 in August 2024, driven by the new dynamics surrounding immigration policy for UK care worker visas. The Office for Budget Responsibility\footnote{https://obr.uk/} projects that UK-wide public spending on adult social care would need to increase by 3.1\% annually in real terms over the next decade.

Globally, the demographic shifts and increasing longevity are also contributing to the evolving needs of the ageing population worldwide \cite{WorldEconomic2023Report1}. By 2050, the global population of individuals aged 60 and above is projected to double from 1 billion to 2.1 billion, with the number of people aged over 80 expected to triple, rising to 426 million. This demographic shift brings with it a surge in demand for comprehensive care services, as an increasing number of older adults require assistance with daily living activities and management of chronic health conditions. Notably, the majority of older adults (up to 80\% by 2050) will reside in low- and middle-income countries, where healthcare infrastructure and social support systems often lag behind those in developed economies \cite{WorldEconomic2023Report1}. 

According to a recent report by WHO \cite{WHOReport}, less than 60\% of countries have integrated long-term care into their national frameworks for geriatric care, leaving significant gaps in service provision. This care crisis is further exacerbated by workforce shortages and the rising number of older adults living alone, particularly in regions with shrinking family structures. As a result, millions of seniors experience unmet care needs, affecting their independence and quality of life.

\begin{figure} [ht!]
\begin{center}
\includegraphics[width=1\columnwidth]{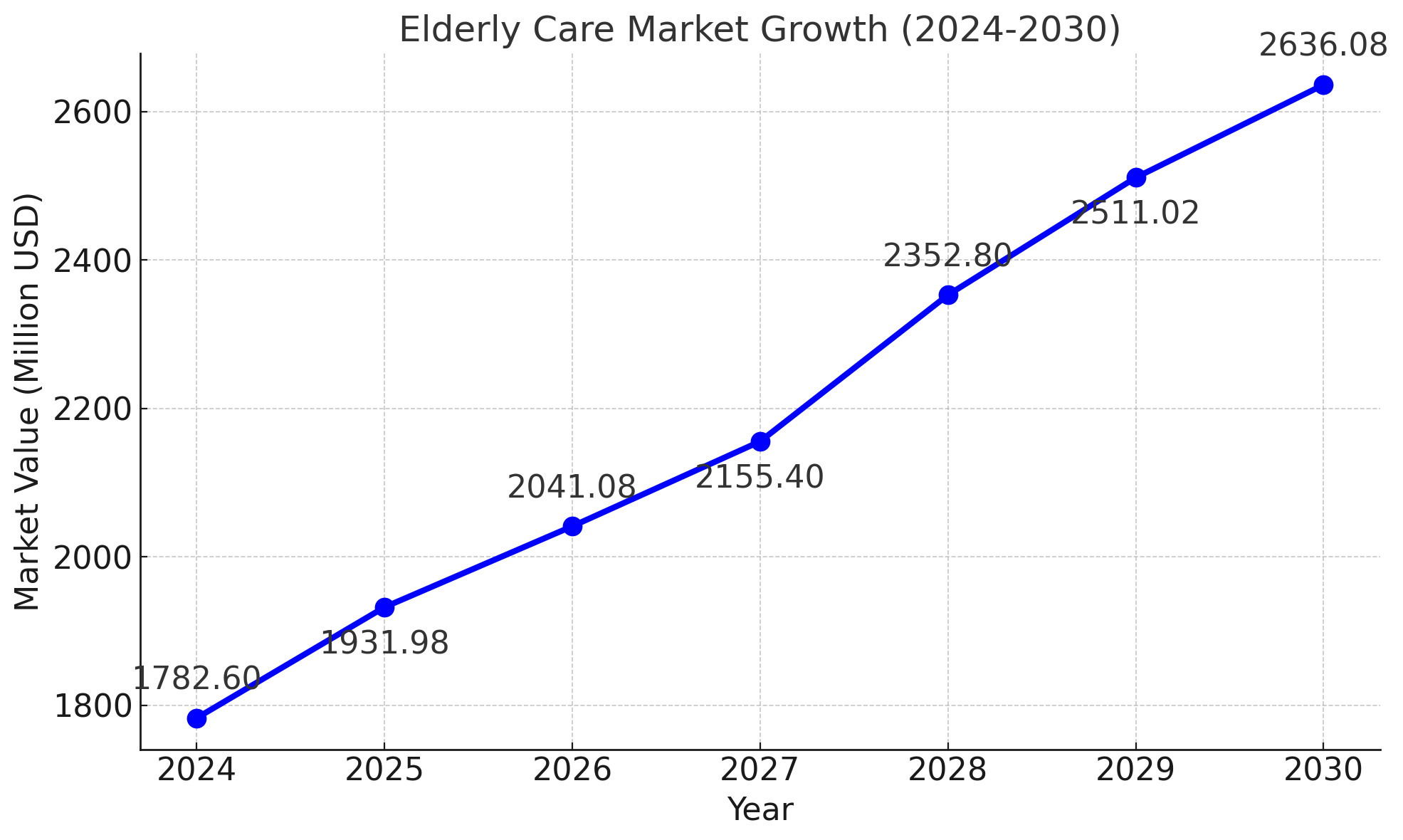} 
\caption{Projected growth of elderly care trends by 2030.}\label{markettrends}
\end{center}  
\end{figure}

These growing challenges and concerns emphasize the urgent need for innovative AI-driven solutions that can autonomously address pervasive issues such as loneliness, cognitive decline, and complex health management in the elderly population \cite{soubutts2023aging}.
The growing interest in AI for elderly care globally reflects an increasing trend toward addressing elderly care challenges through AI-powered solutions. For example, the global elder care assistive robots market is projected to grow at a robust compound annual growth rate (CAGR) of approximately 14.8\% from 2024 to 2030 \cite{grandview2024}. Similarly, the AI-powered solutions segment within elderly care is expected to achieve a CAGR of approximately 9.73\% from 2025 to 2030, reaching a market value of around USD 2250 million by 2030 \cite{accelirate2025}. This has resulted in a market expansion from an estimated USD 1782.60 million in 2024 to an anticipated USD 2636.08 million by 2030, as shown in Fig. \ref{markettrends}. This strong growth in assistive robots is propelled by caregiver shortages and increasing investments in AI-powered robotic aides offering mobility assistance, medication reminders, companionship, and cognitive support to the elderly. Furthermore, these trends reflect the rising demand for personalized, AI-enabled solutions for elderly care, to support the complex health needs of the ageing population.

Addressing these evolving needs requires a fundamental transformation of care systems, moving towards integrated, person-centered, and accessible solutions. The World Health Organization (WHO) calls for a radical shift in how societies value and deliver elderly care, emphasizing the importance of seamless integration between health and social services, support for informal caregivers, and accountability at all levels of governance \cite{WHOReport22}. While previous technological interventions, such as telehealth services and assistive devices, have attempted to address these challenges, they often lack the much-needed personalization and adaptability \cite{retiwalla2024AgenticAI}. As a response, agentic Artificial Intelligence (AI), powered by Large Language Models (LLMs), are being explored as viable tools to mitigate the gaps by providing personalized health monitoring, cognitive support, and companionship. Agentic AI can provide personalized companionship, cognitive stimulation, and health monitoring, addressing these issues more effectively than traditional technologies. Unlike earlier interventions, Agentic AI possesses the capacity to learn, adapt, and make autonomous decisions, offering a paradigm shift in the way care is conceptualized and delivered. These systems can offer tailored support, from managing daily activities to providing cognitive engagement and health monitoring, potentially revolutionizing the approach to caring for an ageing population. Nonetheless, the successful adoption of Agentic AI must be underpinned by principles of equity, inclusivity, and responsiveness to the diverse needs and vulnerabilities of ageing populations.

\subsection{The Agentic AI Revolution and LLMs}
Agentic AI revolutionizes elderly care by combining autonomy with advanced language capabilities, enabling meaningful engagement and dynamic adaptability. Unlike traditional AI, which follows predefined instructions, Agentic AI acts proactively to achieve specific goals. For instance, an Agentic AI model can tailor its interactions to an individual's communication style while engaging in human-like conversations, fostering a personalized and empathetic relationship. Additionally, Agentic AI improves informational trustworthiness through advanced cognitive reasoning, reducing the likelihood of errors and ensuring that decisions are based on reliable data sources. 

According to Taha et al. \cite{agenticAI2024taha}, by 2025, global data generation is expected to exceed 180 zettabytes, with healthcare contributing over one-third. Currently, only 3\% of this data is effectively utilized, often due to system inefficiencies. A significant advantage of Agentic AI lies in its ability to process and seamlessly integrate vast amounts of healthcare data, including data from wearable devices, electronic medical records, and smart home systems. For example, it can monitor health metrics in real-time, flag critical changes, and coordinate care plans, providing caregivers with actionable insights while enhancing patient well-being. Agentic AI's chaining capabilities allow it to break down complex problems into actionable steps, offering scalable solutions in elderly care \cite{bernard2024agenticAI}. For instance, it can autonomously manage tasks like medication reminders, adjust communication styles for individuals with cognitive impairments, and notify caregivers of emerging health risks. Notably, Agentic AI may streamline care procedures and worksflows by automating scheduling and triaging high-risk cases, thus, potentially reducing missed care rates, as seen in oncology studies \cite{Tanveer2025Report}.

However, as Agentic AI systems grow more autonomous, the associated ethical challenges, such as accountability, privacy, and safety also grow. Robust measures, including human-in-the-loop mechanisms and compliance with standards like HIPAA and GDPR, have to be in place to ensure reliability. With such safeguards, Agentic AI empowered by LLMs has the potential to redefine elderly care by offering a blend of personalized engagement, proactive healthcare management, and emotional support, enhancing both independence and quality of life for older adults. 

Although Agentic AI is not a new concept, its recent rise in popularity can be attributed to the rapid development of LLMs, which have significantly enhanced its capabilities, particularly in healthcare \cite{bohr2020rise}. These advancements allow Agentic AI systems to function as intelligent conversational agents, capable of understanding complex instructions and engaging in natural conversations. For example, a McKinsey report indicates that AI technologies, including agentic systems, could potentially create \$2.6 trillion to \$6.2 trillion in value annually in 16 business functions, including healthcare, by optimizing operations or improving patient outcomes.

Modern LLM-based platforms\footnote{Few popular examples include ChatGPT, Gemini, Claude, DeepSeek, Grok. The provision of a complete listing is not the objective of this text.} exemplify these capabilities facilitating real-time responsiveness, adaptiveness to individual user needs and context, nuanced comprehension, and task automation. Additionally, models like Anthropic's Claude and Meta's LLaMA are designed to understand nuanced queries and engage in more sophisticated dialogues. Furthermore, recent surveys indicate growing confidence among healthcare professionals that AI will significantly enhance their ability to provide patient care within the next five years \cite{micheal2020generativeAI}. This adaptability and responsiveness make Agentic AI an invaluable tool for improving user experiences and operational efficiency across various applications, particularly in the care sector, where timely and accurate information is crucial.

\subsection{Scope of the article}
In this article, we explore the potential of LLM-based Agentic AI to revolutionize elderly care, examine its applications, challenges, and highlight ethical implications. Specifically, we discuss the role of Agentic AI as autonomous caregivers, meeting the complex needs of older adults through personalized interactions, health monitoring, and cognitive support. The discussion encompasses both the technical aspects of LLM-based Agentic AI, including recent advancements in model fine-tuning and multi-modal integration, as well as their practical applications in enhancing elderly care. Our analysis also extends to the economic and workforce landscape of the elderly care industry, highlighting how Agentic AI can help mitigate challenges such as staffing shortages and operational inefficiencies. Through interdisciplinary insights and emerging evidence, we propose a balanced approach that aligns technological innovation with the specific needs and vulnerabilities of the elderly, while addressing critical ethical concerns such as data privacy, algorithmic bias, and equitable access to AI-driven care solutions.

\subsection{Key Contributions}
To the best of our knowledge, this is the first ever study that presents comprehensive interdisciplinary overview of the role of Agentic AI in elderly care. The key contributions of this work are summarized as follows:
\begin{itemize}
    \item We present a detailed overview of LLM-based Agentic AI and its potential applications in elderly care, highlighting its autonomous and proactive decision-making capabilities for personalized health management, cognitive support, and emotional companionship for older adults.
    \item The adoption of Agentic AI in elderly care brings challenges for all the stakeholders, Hence, we identify and discuss key challenges and their potential solutions in the integration of Agentic AI into elderly care settings. 
    \item We share key insights and lessons learned from implementing LLM-based Agentic AI in elderly care, emphasizing the need for ethical safeguards, privacy protections, and transparent decision-making.
    \item Lastly, we address a gap in current literature by analyzing the unique capabilities, applications, and limitations of LLM-based Agentic AI in elderly care, distinguishing it from other AI technologies. 
    \item We provide an interactive companion dashboard for readers, with the intention of updating it with additional scholarly resources on agentic AI in elderly care. The dashboard is accessible at \url{https://hazratali.github.io/agenticai/}
\end{itemize}

\subsection{Related Reviews}
Recent years have seen a surge in reviews exploring the intersection of AI and elderly care, reflecting the growing interest in leveraging advanced technologies to address the complex needs of ageing populations. For instance, Ma et al. \cite{ma2023artificial} discuss the types of AI technologies employed in elderly healthcare, such as machine learning (ML), natural language processing (NLP), and robotics, as well as their applications in health monitoring, smart homes, and therapeutic interventions. The survey also highlights the potential of these technologies to enhance operational efficiency and the associated challenges, such as digital literacy and ethical concerns, providing a comprehensive mapping of the AI landscape in elderly care. Loveys et al. \cite{loveys2022artificial} discuss the acceptability and effectiveness of AI-enhanced interventions for older people receiving long-term care services. It synthesizes evidence on interventions using social robots, environmental sensors, and wearable devices, revealing mixed outcomes across different health domains and emphasizing the need for more rigorous evaluation of user experience and clinical impact.

Exploring the human side, Wong et al. \cite{wong2025exploring} discuss older adults’ perspectives and acceptance of AI-driven health technologies. This survey investigates key aspects, including attitudes, facilitators, and barriers to adoption, such as perceived usefulness, ease of use, privacy concerns, and the importance of user-friendly design. Using frameworks like the COM-B model, the study provides actionable strategies for improving technology uptake among seniors. Additionally, in a policy-focused review, Zhao et al. \cite{zhao2024opportunities} discuss the integration of AI into social elderly care services, particularly in China. The review analyzes the current landscape, identifies key challenges in AI integration, and proposes policy recommendations to support efficient and equitable care delivery in the face of demographic shifts and resource constraints.

Despite the breadth of existing surveys, a notable gap remains in the literature regarding the specific role of Agentic AI in elderly care. Most prior reviews have treated AI as a broad field, without distinguishing the unique capabilities and challenges of Agentic AI - such as autonomous reasoning, proactive decision-making, and multi-agent collaboration. Our work addresses this gap by offerig a focused analysis of how Agentic AI powered by LLMs can redefine elderly care across domains like personalized health management, cognitive support, and companionship. We critically examine the ethical, technical, and practical challenges specific to Agentic AI and propose a framework for their responsible integration that prioritizes human-centered design and safeguards for vulnerable populations. This targeted approach not only bridges a crucial gap in the literature but also lays the groundwork for the safe and effective adoption of Agentic AI in the care of ageing populations. Table \ref{tab:related-surveys} summarizes key differences between prominent related surveys and the present work.

\renewcommand{\arraystretch}{1.3}
\begin{table*}[ht!]
\centering
\caption{Comparison of the presented work with related review articles.}
\begin{tabularx}{\textwidth}{|p{2.1cm}|p{1.65cm}|p{2.5cm}|p{4.0cm}|X|}
\hline
\hline
\textbf{Survey/Review Title} & \textbf{Focus Area} & \textbf{AI/LLM/Agentic AI Scope} & \textbf{Challenges Addressed} & \textbf{Key Contributions} \\
\hline
\hline
AI in Elderly Healthcare: Scoping Review (2023) \cite{ma2023artificial} & General AI applications in elderly care & Broad AI, not LLM-specific, does not cover Agentic AI & 
\begin{itemize}
    \item Data interoperability issues
    \item Low digital literacy among elderly
    \item Privacy and ethical concerns
\end{itemize} & 
\begin{itemize}
    \item Cataloged AI tools in elderly care
    \item Highlighted ethical and practical barriers
    \item Provided overview of smart homes and health monitoring
\end{itemize} \\
\hline
AI-Enhanced Interventions in Long-Term Care (2022) \cite{loveys2022artificial} & Effectiveness of AI in long-term care settings & Robotics, sensors, basic ML, does not cover LLMs or Agentic AI & 
\begin{itemize}
    \item Workflow integration challenges
    \item Sensor/data reliability limitations
    \item Difficulty measuring outcomes
\end{itemize} & 
\begin{itemize}
    \item Synthesized evidence on AI interventions in long-term care
    \item Evaluated user experience
    \item Identified gaps in clinical outcomes research
\end{itemize} \\
\hline
Older Adults' Acceptance of AI Technologies (2025) \cite{wong2025exploring} & User perspectives and adoption factors & General AI, not LLMs or Agentic AI & 
\begin{itemize}
    \item Usability and trust issues
    \item Cognitive and interaction barriers
    \item Concerns over privacy and autonomy
\end{itemize} & 
\begin{itemize}
    \item Explored older adults’ attitudes toward AI
    \item Identified adoption barriers and facilitators
    \item Used behavioral models for analysis
\end{itemize} \\
\hline
AI in Social Elderly Care Services (Policy Analysis) (2024) \cite{zhao2024opportunities} & Regional integration of AI in social care & General AI, not LLMs or Agentic AI & 
\begin{itemize}
    \item Policy and regulatory gaps
    \item Resource and equity constraints
    \item Governance and accountability issues
\end{itemize} & 
\begin{itemize}
    \item Analyzed policy frameworks for AI in social care
    \item Provided region-specific recommendations
    \item Addressed resource and governance issues
\end{itemize} \\
\hline
Redefining Elderly Care with Agentic AI \textbf{(Ours)} & Agentic AI for elderly care transformation & Agentic AI (autonomous, proactive, multi-agent), focus on LLM-based Agentic AI & 
\begin{itemize}
    \item Reliability and misinformation risks
    \item Data privacy, bias, and fairness
    \item Explainability and regulatory compliance
\end{itemize} & 
\begin{itemize}
    \item Focuses exclusively on Agentic AI and their unique capabilities
    \item Critically examines proactive, autonomous, and multi-agent LLM applications
    \item Addresses specific ethical, technical, and adoption challenges of LLMs
    \item Proposes a comprehensive, human-centered framework for responsible integration
    \item Bridges the gap in literature by offering a forward-looking, LLM-centric perspective
\end{itemize} \\
\hline
\hline
\end{tabularx}
\label{tab:related-surveys}
\end{table*}

\begin{figure*}[ht!]
\centering
\includegraphics[width=0.95\textwidth]{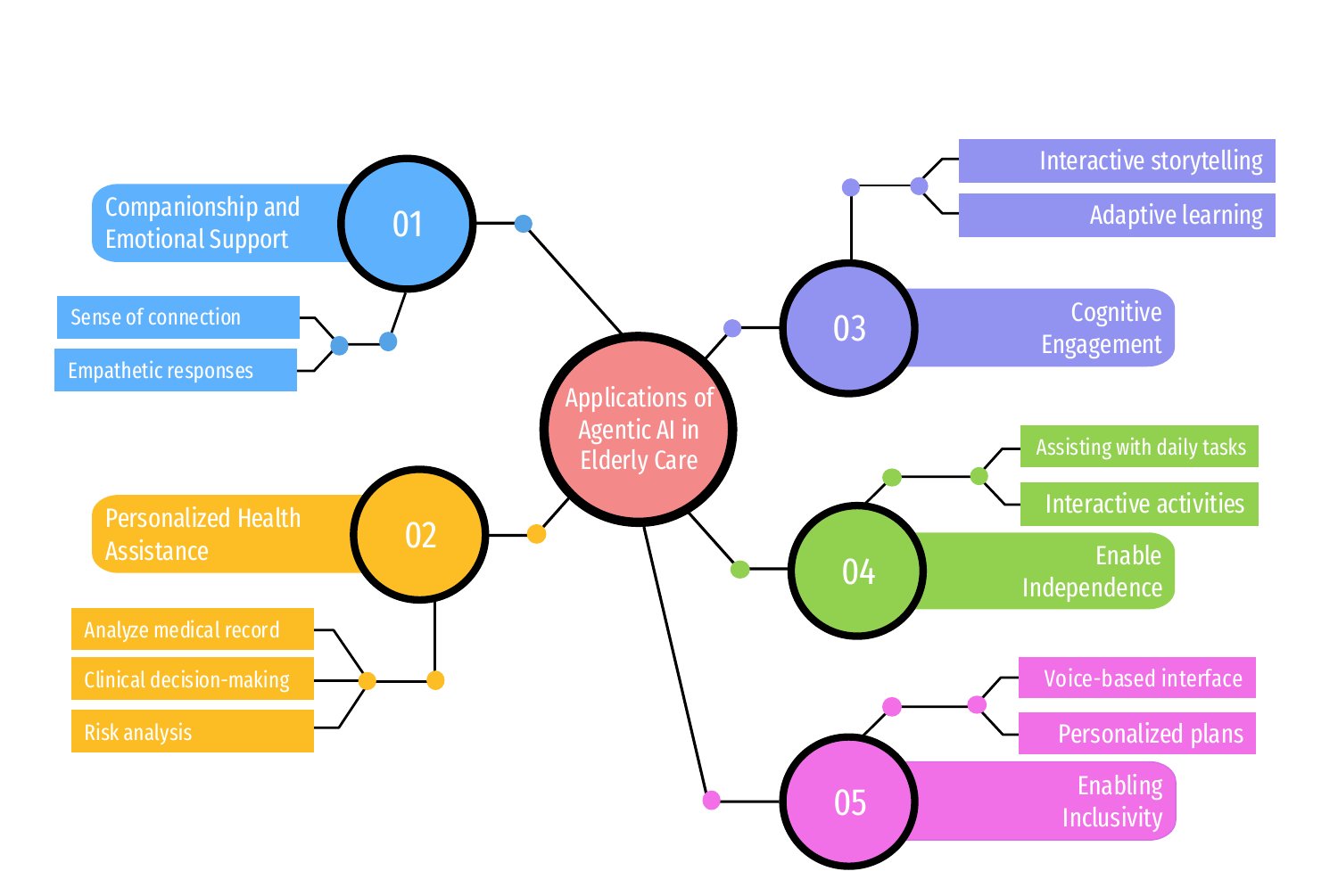}
\caption{Applications of Agentic AI in Elderly Care. The applications are organized into four broad categories in the context of activity. (i) Enabling independence, (ii) Companionship, (iii) Assistance in healthcare decision-making, (iv) Cognitive engagement, (v) Enabling Inclusivity.\label{fig_applications_of_agentic_llms}}
\end{figure*}

\section{Applications of Agentic AI in Elderly Care}
Agentic AI is emerging as a transformative solution to the multifaceted challenges of elderly care, offering personalized, autonomous decision-making capabilities to enhance the quality of life for older adults. Advanced LLM-based chatbots utilize transformer architectures and self-attention mechanisms to generate contextually relevant, human-like language \cite{AIhelper2024humeAI}. Beyond basic conversational interfaces, LLM-based Agentic AI can be fine-tuned for specialized tasks using techniques such as incremental pre-training (IPT) and supervised fine-tuning (SFT), enhancing task-specific performance in elderly care scenarios. For example, Sun et al. \cite{sun-etal-2025-enhancing} demonstrated the fine-tuning of LLMs with IPT and SFT, achieving impressive precision and F1 scores of 86.78\% and 86.21\%, respectively, in elderly care tasks. Fig. \ref{fig_applications_of_agentic_llms} presents a comprehensive overview of the applications of Agentic AI in elderly care.

The integration of Agentic AI in elderly care focuses on enabling independence and fostering cognitive engagement \cite{abadir2024enhancing}. Such systems can help older people schedule their routine appointments, manage smart home controls, and access online services, thus, offering greater autonomy in activities of daily living, especially for individuals requiring routine assistance \cite{khalil2019speech}. Agentic AI can effectively engage older adults in interactive activities, such as storytelling, quizzes, and cognitive exercises, contributing to sustained cognitive resilience in ageing populations. For example, LLM-powered virtual assistants can deliver personalized support, offer cognitive enrichment, and even predict emotional shifts based on language patterns, which can help in tailoring care plans to individual needs \cite{roberts2024supporting}. These capabilities enable the development of AI-driven nursing assistants capable of real-time patient monitoring, personalized interventions, and specialized task execution.

The potential of Agentic AI in elderly care is further exemplified by its ability to provide personalized care insights, assist in clinical decision support, and contribute to predictive analytics in healthcare management for older adults \cite{momand2025integrating}. Building on these capabilities, recent frameworks with retrieval-augmented generation (RAG) combined with foundation model fine-tuning has shown promising results in developing Caregiving Language Models (CaLMs) tailored for dementia care \cite{parmanto2024reliable}. 

In the following subsections, we highlight the potential applications of Agentic AI in elderly care.

\subsection{Companionship and Emotional Support}
In older age, the loss of partners and friends and reduced mobility are common challenges that badly affect their mental and physical health. In such cases, companionship and emotional support are critical for maintaining their mental and emotional well-being. Recently, LLMs have been used to combat loneliness and social isolation among seniors by functioning as conversational agents. OpenAI's ChatGPT and Google's LaMDA are prime examples of AI tools that can engage in meaningful conversations, share stories, and even play interactive games \cite{fear2023shaping}. These systems employ advanced natural language understanding techniques to recall past interactions, adjust their tone and style to the user's preferences, and offer consistent emotional support. For instance, in \cite{armbruster2024doctor}, ChatGPT has offered empathetic responses that are often rated higher in quality than those provided by human physicians.

One notable application is the use of Agentic AI for personalized reminiscence therapy. By integrating with datasets containing personal history or preferences, models like LLaMA can curate tailored content such as music, photos, or stories from a senior's past. This approach not only improves mood but also enhances communication skills and fosters a sense of connection \cite{sun-etal-2025-enhancing}. Additionally, AI-powered systems built on top of LangChain architecture have been deployed to create interactive workshops where seniors participate in group activities led by virtual assistants. These workshops stimulate cognitive function while simultaneously fostering social engagement \cite{duan2024elderqa}.

\subsection{Personalized Health Assistance}
Agentic AI is emerging as a pivotal tool in providing personalized health assistance. Models like ChatGPT and GLM4, when fine-tuned for medical applications, can analyze medical records, monitor vital signs through wearables, and provide real-time insights into health markers  \cite{techtarget2024LLMs}. For instance, ChatGPT has been utilized to support clinical decision-making around medication management for older adults. Similarly, the effectiveness of GLM4 has been demonstrated in patient monitoring tasks by leveraging supervised fine-tuning (SFT) techniques tailored for elderly care \cite{digital2024LLMs}.

LLMs can empower Agentic AI to excel at predictive healthcare management. AlayaCare's LLM-based smart assistant integrates predictive algorithms to identify patients at risk of hospitalization based on clinical data such as comorbidities or previous falls \cite{sun-etal-2025-enhancing}. This proactive approach enables caregivers to intervene early, reducing adverse events such as emergency room visits or preventing avoidable hospitalizations. Furthermore, dietary recommendations powered by LLMs can be customized for conditions like diabetes or hypertension by analyzing an elderly individual's health profile and suggesting tailored nutritional guidance, such as recommending a DASH\footnote{DASH: Dietary Approaches to Stop Hypertension} diet plan or maintaining the glycemic index scoring \cite{fear2023shaping}. These capabilities demonstrate how Agentic AI can enhance preventive care while reducing the burden on healthcare providers.

\subsection{Cognitive Engagement}
LLMs are enhancing cognitive stimulation for the elderly through personalized activities designed to maintain mental sharpness. Models like OpenAI's ChatGPT have been used to create interactive storytelling sessions where seniors contribute to narratives or solve puzzles collaboratively \cite{fear2023shaping}. These activities target specific cognitive functions, such as memory retrieval and problem-solving, while offering entertainment. Additionally, the LLM-based agents have been employed to develop adaptive learning modules that adjust difficulty levels based on the user's cognitive performance \cite{sun-etal-2025-enhancing}.

Another innovative application is the use of Agentic AI tools in enabling context-aware social engagement, for example, through virtual book clubs or discussion groups. For example, Meta’s LLaMA can facilitate discussions about literature or current events, tailoring the complexity of conversations to match the participant's cognitive abilities. This not only stimulates intellectual engagement but also fosters social interaction among group members \cite{treder2024introduction}. Moreover, Agentic AI can assist in reminiscence therapy by curating multimedia content from a senior's life history, such as photos, music, and personal autobiography notes, helping them reconnect with memories and achieve delayed cognitive decline. 

\subsection{Enabling Independence}
Agentic AI is empowering seniors to maintain independence by assisting with daily tasks and decision-making. Multi-modal AI systems integrate voice, text, and visual inputs to create intuitive interfaces that cater to diverse needs \cite{scottcode2024LLMs}. For example, these systems can help seniors schedule appointments, manage medication reminders, or navigate online services with minimal assistance. A notable development is the integration of Agentic AI with smart home technologies. By connecting with internet of things (IoT) devices, platforms such as AlayaCare's smart assistant enable dynamic environmental control, such as controlling lighting or temperature based on real-time preferences and behavioral cues \cite{newoAI2024}.
Additionally, these systems can proactively identify needs, for instance, ordering groceries when supplies are low, or guide seniors through complex tasks, such as online banking, using step-by-step instructions. Such applications not only enhance day-to-day autonomy but also serve to narrow the digital divide, empowering older adults to engage confidently with technology. For example, if a senior experience discomfort due to room temperature, the Agentic AI can automatically trigger temperature adjustments, offering comfort and convenience. 

The agentic capabilities of Agentic AI in elderly care go beyond environmental control, encompassing proactive health monitoring and decision-making. These systems can analyze data from wearable devices and home sensors to detect potential health anomalies, automatically scheduling medical appointments or alerting caregivers when unusual patterns are detected \cite{jin2022synapse}. Services like SeaX Voice AI demonstrate how AI can conduct thousands of automated check-in calls, monitor seniors' well-being, and provide immediate follow-up when responses are received. Furthermore, Agentic AI are revolutionizing cognitive engagement for seniors by creating personalized mental stimulation activities tailored to individual cognitive abilities and interests. These AI systems can design interactive storytelling sessions, memory games, and adaptive learning experiences that help maintain mental acuity and potentially slow cognitive decline \cite{koebel2021expert}. By remembering past interactions and preferences, Agentic AI can create increasingly sophisticated and meaningful engagement strategies that not only provide mental stimulation but also offer emotional support. As these technologies continue to evolve, they promise to enhance the independence, safety, and overall quality of life for older adults, addressing critical challenges in elderly care such as social isolation, healthcare management, and maintaining cognitive function.

\subsection{Enable Inclusivity}
AI voice agents, powered by LLMs, are poised to revolutionize elderly care by enabling natural, real-time conversational interactions that extend the reach and capacity of healthcare systems \cite{adams2025generative}, thus, enabling inclusivity. Unlike traditional scripted chatbots, these agents generate context-sensitive, personalized responses by integrating extensive medical knowledge, patient histories, and dynamic task lists, allowing them to handle complex clinical nuances and unexpected questions with natural speech \cite{Car2020}. Such capabilities are particularly well-suited for the elderly, many of whom encounter barriers to digital technology and find promise in voice-based interfaces that lower the threshold for engagement, and offer a more \textbf{accessible means of interaction} while providing \textbf{companionship}, \textbf{health guidance}, and \textbf{administrative assistance}. 

A promising application of AI voice agents in elderly care is chronic disease management and early symptom triage. These agents conduct regular check-ins to monitor changes in symptoms or mood, enabling timely detection of clinical deterioration and reducing unnecessary hospitalizations \cite{Bhimani2025}. In addition, they support medication adherence, provide tailored health education, and escalate urgent concerns to clinicians when needed. On the operational side, AI voice agents automate appointment scheduling, prescription refills, insurance verification, and transportation coordination, alleviating caregiver burden and improving healthcare access \cite{Hyro2025,Orbita2025}. Conclusively, these systems can adapt communication style and language to match patients’ health literacy and cultural background, enhancing inclusivity and reducing disparities in under-served elderly populations \cite{Bhimani2025}.

Many seniors face challenges such as declining vision, hearing loss, reduced mobility, or cognitive impairments, making it essential for AI-driven solutions to offer accessible interfaces and adaptive features. Agentic AI excels in this regard by personalizing care plans, reminders, and health interventions in real time, taking into account each individual’s unique needs and preferences \cite{Edge2025Report}. For example, agentic systems can integrate voice commands, large text displays, and simplified navigation to accommodate users with visual or dexterity limitations, while also providing multilingual support and culturally sensitive content to enhance comfort and acceptance among diverse populations. This adaptability not only improves user experience but also empowers seniors to maintain independence and engage more fully with their care routines.

\begin{figure*}[ht!]
\centering
\includegraphics[width=0.9\textwidth]{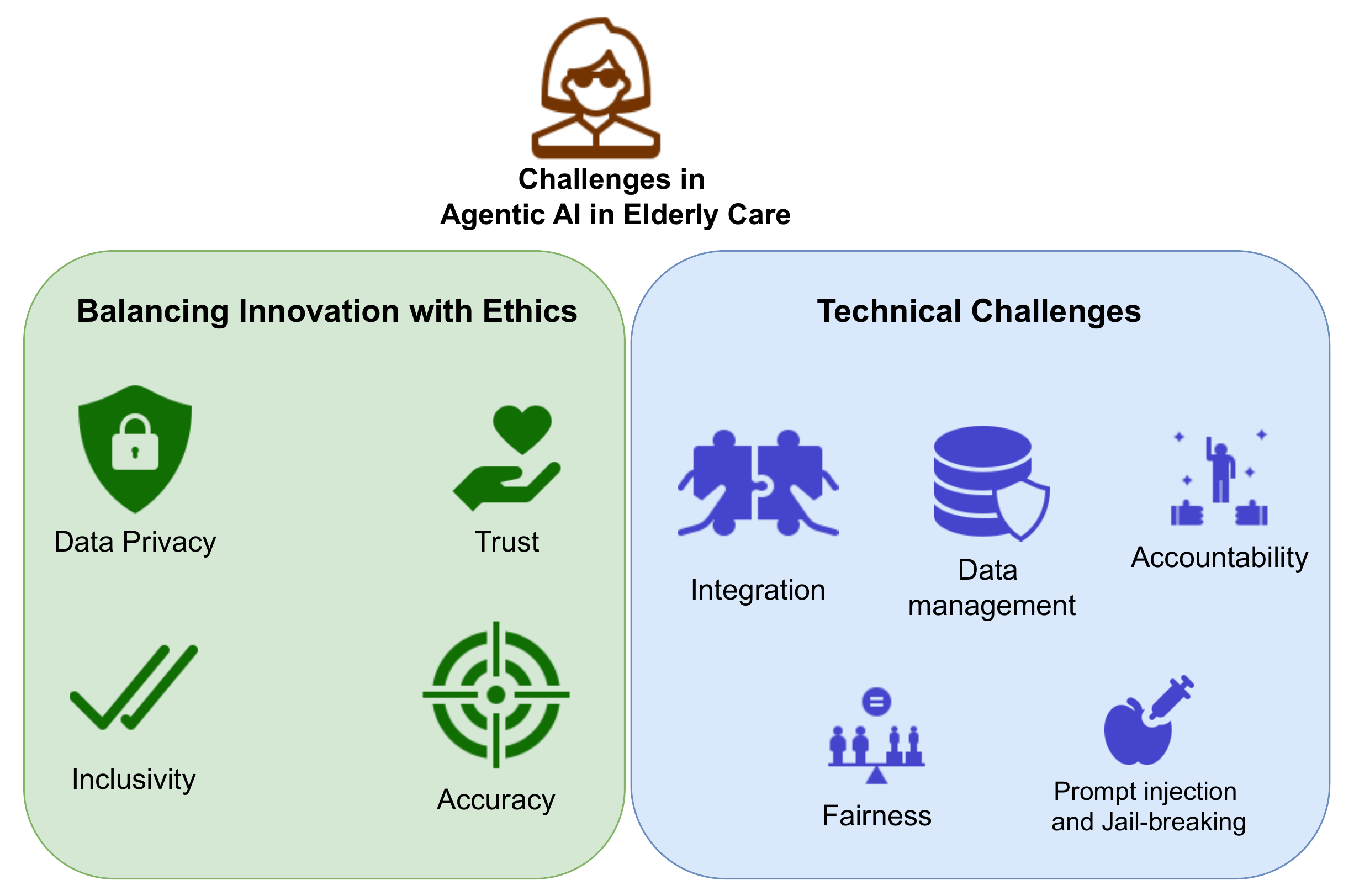}
\caption{Challenges associated with Agentic AI in Elderly Care. The challenges can be broadly categorized as; (i) the challenge of balancing innovation with ethics, and (ii) technical challenges.}\label{fig_challenges}
\end{figure*}

\section{Challenges and Potential Solutions}
In this section, we outline the key challenges that arise when integrating Agentic AI into elderly care, and highlight potential solutions to address these challenges. As the following subsections detail, it is essential to balance rapid technological innovation with strong ethical safeguards, particularly around privacy and security of the elderly healthcare data. Ensuring the accuracy and trustworthiness of AI-generated information is critical to prevent the spread of misinformation, which can have serious consequences in the healthcare settings. Additionally, inclusivity must be prioritized so that solutions are accessible and usable for older adults with diverse abilities/needs and varying levels of digital literacy. By systematically addressing these challenges, we can pave the way for more reliable, equitable, and human-centered deployment of Agentic AI in elderly care.

\subsection{Balancing Innovation with Ethics}
The rapid integration of agentic LLMs into elderly care creates two main priorities. First, there is a need to use technological innovation to improve care. Second, it is crucial to maintain robust ethical standards to safeguard vulnerable populations~\cite{ferri2025multiagent,yuan2025agentic,plivo2025agentic}. Agentic AI systems are becoming increasingly autonomous, as they can make proactive decisions and handle complex tasks with minimal human intervention. This autonomy brings numerous benefits, enhancing operational efficiency, minimizing administrative tasks, and providing personalized support to older adults~\cite{beam2025revolution,yuan2025agentic}. However, greater autonomy also brings more risks, resulting in issues such as data privacy, security, and accountability, which have become increasingly severe. It is crucial to ensure that innovation does not outpace ethical safeguards using Agentic LLMs, which often process sensitive health data ~\cite{statusneo2025privacy,plivo2025agentic}. They connect with many digital systems, which increases the risk of data breaches and unauthorized access. There is also a risk of misinformation where AI-generated outputs may contain errors or hallucinations, which can be particularly hazardous in healthcare settings~\cite{statusneo2025privacy,plivo2025agentic,han2024medical, zhu2024potential}. Fig.~\ref{fig_challenges} categorizes the key challenges associated with Agentic AI in elderly care, distinguishing between the challenge of balancing innovation with ethics and the technical challenges such as accuracy, inclusivity, and data management.

To address these challenges, a multi-layered approach is needed, utilizing technical safeguards such as differential privacy and federated learning, which are essential~\cite{ferri2025multiagent,yuan2025agentic,accelirate2025Blog,ferri2025multiagent,yuan2025agentic}. Transparent decision-making and human-in-the-loop validation help build trust and ensure safety. Inclusivity is also a vital aspect of AI solutions that must be accessible and equitable for all older adults, regardless of their abilities or digital skills~\cite{bhamidipaty2025revolutionizing,ufl2024ageism}. By focusing on both ethics and technology, agentic LLMs can responsibly transform elderly care. This approach helps protect the dignity, independence, and well-being of the aging population~\cite{beam2025revolution,yuan2025agentic}.

\subsubsection{Data Privacy and Security}
Balancing innovation with ethics, data privacy, and security is one of the key challenges 
in elderly care \cite{David2025Report, Katie2025Report}. Agentic AI-based applications are designed to act autonomously, making decisions and executing tasks with minimal human intervention. 
However, this increased autonomy also introduces new risks related to data privacy and security. Agentic AI often processes sensitive patient data and interacts with multiple external systems, expanding the potential attack surface for Cyberthreats. The shift to Agentic AI can streamline a healthcare call center's operations, but only if robust safeguards are in place to prevent data breaches and unauthorized access \cite{Tanveer2025Report}. 

In this regard, various techniques such as differential privacy, federated learning (FL), and secure multi-party computation are being explored to enhance data protection, while metrics like mean time to detect (MTTD) and mean time to respond (MTTR) to security incidents are becoming benchmarks for evaluating system resilience \cite{Dwayne2025blog}. Ethically, the deployment of Agentic AI requires transparent decision-making processes and clear accountability. According to Andy et al. \cite{Andy2025Report}, the risk of misinformation remains significant, with recent estimates placing LLM hallucination rates between 5\% and 30\% as of March 2025. This underscores the need for multi-agent consensus mechanisms and human-in-the-loop validation to ensure the accuracy and trustworthiness of AI-generated outputs.

Furthermore, the financial implications of Agentic AI adoption are substantial. Consumption-based pricing models for healthcare AI agents range from \$4.02 to \$5.99 per hour or successful outcome, with the cost of building digital teammates estimated between \$500,000 and \$1 million per use case \cite{Tanveer2025Report}. These investments can yield significant returns by optimizing resource allocation and improving patient satisfaction. Also, they can necessitate rigorous oversight to ensure that cost savings do not come at the expense of privacy or ethical standards. The integration of Agentic AI in elderly care presents both transformative opportunities and complex challenges. By implementing advanced technical safeguards, adopting transparent and accountable governance frameworks, and continuously monitoring key metrics such as hallucination rates \cite{asgari2025framework}, security incident response times, and user satisfaction, healthcare organizations can harness the benefits of Agentic AI while upholding the highest standards of ethics, privacy, and security.

\subsubsection{Accuracy and Trustworthiness}
Accuracy and trustworthiness are critical concerns in the deployment of Agentic AI for elderly care, as LLM-based tools are prone to generating misinformation, commonly referred to as hallucinations. Misinformation in LLMs can manifest in several forms, including fabricated facts, misleading claims, out-of-context information, and even biased or fabricated citations \cite{ali2023chatgpt,Misinformation2025Report}. Alarmingly, recent analyses estimate hallucination rates for LLMs to range between 5\% and 30\% as of early 2025 \cite{Andy2025Report}. In healthcare settings, such errors can have severe consequences. For instance, if an LLM provides incorrect dosage information for medication or misinterprets a user's symptoms, it could result in harm or delayed medical care \cite{andrew2024potential}. This underscores the need for multi-agent consensus mechanisms and \textbf{human-in-the-loop} validation to ensure the accuracy and trustworthiness of AI-generated outputs. The challenge is compounded by the fact that LLMs often present both accurate and erroneous content with similar levels of confidence and fluency, making it difficult for users, especially older adults who may lack technical expertise, to discern truth from error \cite{chen2024combating}. Furthermore, the potential misuse of Agentic AI by malicious actors to intentionally generate plausible-sounding but factually incorrect medical information further undermines public trust and increases the risk to vulnerable populations. Similarly, user acceptance and trust critically influence the sustained use of generative AI voice agents by older adults. 

\subsubsection{Inclusivity}
Inclusivity is a cornerstone in the design and deployment of Agentic AI for elderly care, ensuring that technological advancements benefit all older adults regardless of their abilities or digital literacy levels \cite{Vines2015Ageing}. However, bridging the digital divide requires more than just developing accessible technology; it demands robust training, ongoing support, and active community engagement. Many elderly individuals may be unfamiliar or uncomfortable with new digital tools, so offering step-by-step tutorials, responsive help desks, and caregiver involvement is vital for fostering confidence and sustained use \cite{Jesse2025Report}. Technical solutions must incorporate adaptive interaction modalities, such as voice recognition, gesture control, and simplified graphical user interfaces, to accommodate users with visual, auditory, or motor impairments \cite{Lexie2019Older,stein2021accessible}. Furthermore, AI systems should employ personalization algorithms that dynamically adjust content delivery and interaction complexity based on the user's abilities and preferences \cite{rashidi2012survey}. Equally important are robust training programs and ongoing technical support. Studies highlight that step-by-step tutorials, context-aware help systems, and the involvement of caregivers or family members can significantly enhance technology adoption and sustained engagement among the elderly \cite{Jesse2025Report,Mitzner2019TechnologySupport}. Community engagement initiatives, such as peer-led digital literacy workshops and participatory design sessions, further empower older adults by incorporating their feedback into system refinement.

\subsubsection{Potential solutions to balance innovations}
The integration of Agentic AI in elderly care requires robust technical safeguards to protect sensitive health data. Implementing end-to-end encryption combined with federated learning architectures ensures data privacy while maintaining model performance \cite{EdgeAI2025}. Multi-agent consensus mechanisms coupled with human-in-the-loop validation can reduce hallucination rates below 5\% by cross-verifying outputs through specialized sub-agents like Validator and Critic models \cite{micheal2020generativeAI,adams2025generative}. Adaptive interfaces using contrastive learning approaches enable multilingual support and cultural customization, particularly crucial for diverse ageing populations \cite{accelirate2025,NasirHaslam2025}. Blockchain-integrated logging systems create immutable audit trails for accountability, while real-time bias detection algorithms paired with automated fairness reports help caregivers monitor equity in AI recommendations \cite{Newo2025,kaliappan2024exploring}.
Mitigating the risks of misinformation and hallucinations requires a multi-layered approach. Technically, strategies such as integrating real-time fact-checking, using external and up-to-date data sources, and employing multi-agent consensus mechanisms can help reduce the rate of hallucinations and misinformation \cite{kaliappan2024exploring}. For example, recent research shows that advanced LLMs like GPT-4.5 and Claude 3.7 have lower hallucination rates than earlier models, but no model is entirely immune to error. Prompt engineering, careful configuration settings, and continuous fine-tuning with domain-specific data are also important for improving output reliability \cite{murugesan2025rise}. On the user side, clear disclaimers, transparency about model limitations, and human-in-the-loop validation-where caregivers or clinicians review and approve critical outputs-are essential safeguards. Establishing universal guidelines for the validation and auditability of AI-generated health information, along with regular monitoring of hallucination rates and user feedback, can further enhance trustworthiness and safety in agentic LLM-driven elderly care systems \cite{ali2023chatgpt}. Table \ref{tab:updated-technical-challenges} outlines the main challenges encountered during the integration of Agentic AI in elderly care, along with potential solutions to address these challenges, including issues related to data privacy, system interoperability, and bias mitigation.

To address challenges related to inclusivity, feedback mechanisms and user-centric design approaches allow developers to continuously refine agentic AI solutions based on real-world experiences, ensuring that evolving needs and challenges are promptly addressed. Community partnerships and collaboration with caregivers help tailor solutions to specific cultural contexts and daily realities, further enhancing inclusivity and adoption. By prioritizing accessibility, training, and feedback, Agentic AI can help close the digital gap, reduce social isolation, and ensure that the benefits of AI-powered elderly care are equitably distributed across all segments of the ageing population.

\begin{table*}[ht!]
\centering
\caption{Challenges and Potential Solutions for Agentic AI in Elderly Care.}
\label{tab:updated-technical-challenges}
\begin{tabularx}{\textwidth}{|p{2.2cm}|X|X|X|p{1.6cm}|}
\hline
\hline
\textbf{Challenge} & \textbf{Impact on Elderly Care} & \textbf{Example Scenario} & \textbf{Potential Solutions} & \textbf{References} \\
\hline
\hline
Interoperability with Legacy Systems & Limits care coordination and system adoption across healthcare infrastructure & LLM agent fails to access or update patient EMR due to format mismatch & FHIR/HL7-compliant APIs, automated schema mapping, middle-ware platforms & \cite{accelirate2025Blog,NUAI2025Report} \\
\hline
Data Privacy \& Security & Risk of data breaches and loss of trust in AI-powered systems & Sensitive health data leaked due to unsecured cross-system communication & End-to-end encryption, federated learning, secure multiparty computation, real-time monitoring & \cite{David2025Report,Tanveer2025Report,Dwayne2025blog} \\
\hline
Bias \& Fairness & Perpetuates healthcare inequities by reinforcing biased recommendations & Minorities receive less evidence-based suggestions than majority cohorts & Diverse training datasets, fairness audits, adversarial debiasing, explainable AI & \cite{accelirate2025Blog,Kumarappan2025Report,rashidi2012survey} \\
\hline
System Reliability \& Hallucinations & Reduced trust due to clinical inaccuracies and hallucinated outputs & LLM provides fabricated dosage instructions, misinterprets health queries & Multi-agent validation, domain-specific fine-tuning, clinical safety nets, transparent disclaimers & \cite{Andy2025Report,ali2023chatgpt,micheal2020generativeAI,andrew2024potential,chen2024combating} \\
\hline
Scalability \& Real-Time Responsiveness & Limits use in urgent care conditions due to latency or compute bottlenecks & High delay in fall detection or medication reminders & Lightweight models, edge computing, attention pruning, MobileLLM distillation & \cite{Tanveer2025Report,MobileLLM2024,Jin2022smartphone} \\
\hline
Financial Constraints \& Resource Management & Limits adoption in resource-constrained settings like senior clinics & High upfront costs of building/maintaining LLM agents (\$500K–\$1M per case) & Consumption-based pricing, hybrid cloud-edge infrastructure, agentic workload optimization & \cite{Tanveer2025Report,Dwayne2025blog} \\
\hline
Multi-Agent Coordination & Conflicting outputs from agents cause confusion & Critic model and Health Assistant offer contradictory care suggestions & Hierarchical role-based architecture, consensus checks, agent specialization (Teacher, Validator, Critic) & \cite{micheal2020generativeAI,ali2023chatgpt,ferri2025multiagent} \\
\hline
Prompt Injection \& Jail-breaking Vulnerabilities & Malicious prompts could override safeguards, leak data, or deliver harmful advice & Adversary triggers jail-broken LLM to reveal private medical history or unsafe suggestions & Input sanitization, role-based access, red-teaming tools, audit logging, adversarially robust architectures & \cite{yang2024adversarial,han2024medical,lin2024large,ali2023chatgpt,sun-etal-2025-enhancing} \\
\hline
Human-in-the-Loop Oversight & Autonomous LLMs make critical decisions without user review & Medication or emergency response recommendation issued without caregiver involvement & Override mechanisms, confidence thresholds, human approval checkpoints, audit trails & \cite{David2025Report,Tanveer2025Report,Jesse2025Report} \\
\hline
Adaptive Interface \& Inclusivity & Elderly users with impairments face engagement difficulties & AI interface too complex for voice-impaired or low-digital-literacy users & Voice/gesture/UCR input, adaptive difficulty models, multilingual GUI, progressive disclosure design & \cite{Lexie2019Older,Jesse2025Report,Mitzner2019TechnologySupport,Vines2015Ageing} \\
\hline
Synthetic Data Generation \& Privacy Preservation & Lack of shareable training data stalls personalization while risking privacy breach & LLM cannot access adequate real-world data due to privacy laws & Use of differential privacy, GANs, neural diffusion models to synthesize shareable, anonymized datasets & \cite{ho2020denoising,ali2022spot,ali2023leveraging,GANsynthesis2023} \\
\hline
Ethical \& Regulatory Compliance & Risk of violating user autonomy, legal uncertainties across jurisdictions & LLM operates beyond patient consent; audit trails missing during legal queries & Built-in consent flows, explainable AI, logging via permissioned blockchain, compliance auditing & \cite{Tanveer2025Report,ferri2025,Kumarappan2025Report,Newo2025} \\
\hline
\hline
\end{tabularx}
\vspace{1em}
\end{table*}

\subsection{Technical Challenges}
\subsubsection{Data Inconsistencies and Integration within the Existing Systems} The integration of Agentic AI into elderly care brings a complex set of technical challenges that must be addressed to unlock its true potential. One of the foremost issues is ensuring seamless interoperability with the existing healthcare infrastructure. Many elderly care facilities rely on a fragmented ecosystem composed of electronic medical records (EMRs), diagnostic tools, appointment scheduling platforms, and billing systems, many of which are legacy systems that were not designed with advanced AI integration in mind \cite{accelirate2025Blog}. Furthermore, realizing the full utility of Agentic AI requires adherence to standardized data formats (e.g., HL7, FHIR), and real-time synchronization pipelines. Technical hurdles such as data silos, inconsistent data quality, and system incompatibility can hinder the deployment of Agentic AI and limit its ability to deliver holistic, personalized elderly care \cite{NUAI2025Report}.

\subsubsection{Data Management and Security} Another critical technical challenge is the robust management of data, security, and privacy preservation when deploying LLM-based Agentic AI in elderly care. These models must continuously process and analyze large volumes of sensitive information—including health records, data from wearable devices, and even genetic profiles—to deliver personalized services and timely interventions \cite{accelirate2025Blog}. To meet the stringent requirements of regulations such as HIPAA (U.S.) and GDPR (EU), robust technical measures must be implemented at every stage of data handling. This includes the use of advanced encryption protocols, secure storage architectures, and real-time monitoring systems to detect and prevent unauthorized access \cite{Jesse2025Report}. Additionally, LLM-based Agentic AI should be engineered to facilitate transparent data usage and support dynamic consent management, enabling elderly users and caregivers to control, audit, and revoke data permissions as needed. Addressing these technical aspects is essential to safeguarding privacy, building trust, and ensuring the sustainable integration of Agentic AI in elderly care.

\subsubsection{Bias Mitigation and Fairness} Bias and fairness present additional technical obstacles. Agentic AI systems are only as reliable as the data on which they are trained. If these datasets are unrepresentative or contain historical biases, the resulting models may perpetuate or even amplify inequities in care delivery \cite{accelirate2025Blog}. This is particularly concerning in elderly care, where disparities in health outcomes can be exacerbated by algorithmic bias. Rigorous dataset curation, ongoing bias audits, and the use of fairness metrics are necessary to ensure equitable outcomes. Technical solutions, such as adversarial debiasing and explainable AI, can further help identify and mitigate sources of bias, thereby fostering trust and inclusivity in AI-driven elderly care.

\subsubsection{Accountability and Reliability} The autonomous and adaptive nature of Agentic AI introduces challenges related to oversight, accountability, and system reliability. While these models can make proactive decisions and adapt care plans in real time, there must be clear mechanisms for human oversight and intervention, especially in high-stakes scenarios. Developing transparent logging, audit trails, and fail-safe protocols ensures that caregivers can review, validate, or override AI-driven recommendations when necessary \cite{Kumarappan2025Report}. Additionally, maintaining and updating these complex systems requires specialized technical expertise and ongoing evaluation to ensure that Agentic AI remains accurate, secure, and aligned with evolving clinical standards. Addressing these technical challenges is crucial for safely scaling Agentic AI in elderly care and maximizing its benefits for both patients and providers. For instance, safety concerns for AI voice agents are significant. Patients may treat AI-generated medical advice as definitive, risking harm if urgent conditions are missed. To address this, robust clinical safety mechanisms are essential, including domain-specific training to recognize red flags, uncertainty monitoring, and automatic escalation to human clinicians \cite{adams2025generative,Saenz2023}.

\subsubsection{Challenges in Jail-breaking LLMs: Prompt Injection and Security Risks}
As discussed, LLMs have demonstrated significant potential in transforming elderly care through agentic AI. However, a critical technical vulnerability lies in their susceptibility to prompt injection and jail-breaking attacks. Prompt injection occurs when adversaries manipulate model inputs—either directly via user interaction or indirectly through compromised data sources—to override system instructions and elicit unauthorized or harmful outputs~\cite{ali2023chatgpt,yang2024adversarial}. Jail-breaking is a related phenomenon, where carefully crafted adversarial prompts bypass built-in safety mechanisms, enabling the model to generate responses that violate ethical, privacy, or regulatory guidelines~\cite{Hayler2025,han2024medical}.

In the context of elderly care, these vulnerabilities can have severe consequences. For example, successful prompt injection could lead to the disclosure of sensitive health information, dissemination of misinformation, or execution of unauthorized actions, thereby endangering user safety and eroding trust in AI-driven care~\cite{han2024medical, Hayler2025}. The autonomous and adaptive nature of LLM-based agentic AI exacerbates the challenge, as attackers may exploit conversational interfaces to manipulate medication reminders, alter health advice, or access private user data~\cite{sun-etal-2025-enhancing}. Moreover, distinguishing between legitimate and adversarial inputs is particularly difficult for elderly users with limited digital literacy.

Addressing these weaknesses demands robust input validation, continuous monitoring for adversarial behaviors, and human-in-the-loop oversight to ensure that LLM-driven systems remain safe, reliable, and aligned with ethical standards~\cite{ali2023chatgpt,sun-etal-2025-enhancing}. Recent research emphasizes the need for adversarially robust model architectures, multi-agent consensus mechanisms, and transparent audit trails to mitigate risks associated with prompt injection and jail-breaking~\cite{lin2024large}. As agentic AI becomes more prevalent in elderly care, prioritizing security and resilience against these attacks is essential for safeguarding vulnerable populations.

\subsubsection{Potential Solutions for Technical Challenges}
FHIR/HL7-compliant APIs with automated schema mapping enable seamless integration with legacy EHR systems through middleware solutions \cite{accelirate2025}. 
Hybrid cloud-edge architectures employing lightweight model distillation techniques like MobileLLM optimize computational efficiency without sacrificing accuracy \cite{MobileLLM2024}. Differential privacy-preserved synthetic datasets generated via generative models (Generative Adversarial Networks \cite{ali2023leveraging}, Neural Diffusion Models \cite{ho2020denoising,ali2022spot}) address data scarcity issues while maintaining patient confidentiality \cite{GANsynthesis2023}. Hierarchical agent architectures implement role-based task allocation, separating clinical decision-making from administrative functions \cite{NasirHaslam2025,micheal2020generativeAI}. Quantum-resistant encryption protocols future-proof sensitive health data against emerging cryptographic threats \cite{QuantumEncrypt2024,Kumarappan2025}.

Moreover, a multi-layered defense strategy is essential to mitigate the risks of jail-breaking and prompt injection in LLMs. Key solutions include robust input validation and sanitization to filter out potentially malicious prompts before they reach the model, as well as the use of strict output formatting and semantic filters to detect and block unsafe or unauthorized responses \cite{asgari2025framework}. Defining clear system prompts and enforcing strict adherence to context can help constrain the model's behavior. At the same time, privilege control—such as role-based access and least privilege principles—limits the potential damage from successful attacks. Incorporating human-in-the-loop oversight for high-risk or sensitive operations adds a safeguard, ensuring that critical actions require human approval. Regular adversarial testing, including automated red-teaming and fuzzing, is vital for identifying new vulnerabilities and stress-testing model defenses \cite{korinek2023generative}.

\section{Future Directions and Research Priorities}
The rapid evolution of Agentic AI in elderly care necessitates forward-looking research priorities that address both technical innovation and ethical responsibility \cite{ferri2025}. As these systems become more autonomous and pervasive, it is crucial to establish robust frameworks that ensure safety, inclusivity, and continuous improvement. This section outlines key future directions and research priorities for the responsible integration of Agentic AI in elderly care. Table \ref{tab:future_directions} outlines the key future research directions and priorities for Agentic AI in elderly care, focusing on areas such as the development of standardized guidelines, integration of multi-modal information, addressing ethical concerns, and ensuring adversarial robustness.

\subsection{Development of Standardized Guidelines and Prompt Design}
One of the most immediate research imperatives is the development of standardized prompt design guidelines and system validation protocols tailored specifically for elderly care applications \cite{perez2025}. Current approaches often rely on ad-hoc prompt engineering, which often lacks reproducibility and introduces variability in performance \cite{perez2025,healthtech2025}. Structured and transparent prompt design standards are essential to ensure that Agentic AI generates accurate, safe, and contextually appropriate outputs, especially when used by informal caregivers or older adults without clinical expertise \cite{healthtech2025}. This includes stress-testing under edge cases (e.g., rare diseases, cognitive impairments, atypical language use) and simulating interactions with users from diverse linguistic, cultural, and socioeconomic backgrounds \cite{ma2023}. Validation protocols should incorporate not only technical metrics, such as accuracy and latency, but also user-centered metrics, including perceived trust, usability, and risk perception \cite{Asgari2025}.

Furthermore, comprehensive system validation frameworks are needed to test Agentic AI for accuracy, safety, and appropriateness in various elderly care scenarios. These frameworks should include stress-testing with edge cases (e.g., rare diseases, cognitive impairments, atypical language use), and simulating interactions with users from diverse linguistic, cultural, and socioeconomic backgrounds \cite{ma2023}. Regular updates and revisions to these guidelines will be necessary as both the technology and the regulatory landscape evolve. The goal is to build trust among users and stakeholders by demonstrating that Agentic AI is not only robust but also reliable and safe for vulnerable populations \cite{Asgari2025}.

Finally, ongoing education and training programs are crucial for caregivers and healthcare professionals to interact effectively with these systems, understand their limitations, and recognize situations where human intervention is required \cite{clinician2024}. Standardized training materials, supported by practical examples and case studies, can help bridge the gap between technological innovation and real-world application, ensuring that Agentic AI is used to its full potential while minimizing risks.

\subsection{Integration of Multi-modal Information and Features}
The integration of multi-modal information and features is a critical frontier for Agentic AI in elderly care. Unlike LLMs-based tools for text processing, which rely heavily on text, the development of vision-language models opens frontiers for future developments that can leverage voice, vision, and sensor data to create more intuitive and accessible interfaces \cite{jin2022}. For example, integrating voice recognition and synthesis can make systems more user-friendly for older adults with limited digital literacy or physical impairments. Visual aids, such as diagrams or interactive tutorials, can further enhance comprehension and safety for complex care tasks \cite{mecs2016}.

Multi-modal integration also enables Agentic AI to provide more comprehensive support. By processing data from wearable devices, smart home sensors, and electronic health records, these systems can offer real-time health monitoring, early warning of potential issues, and personalized recommendations \cite{wearable2024}. For instance, an agentic LLM could detect a fall using motion sensors, assess vital signs via wearables, and immediately notify caregivers or emergency services. Such capabilities can significantly improve the independence and safety of older adults living alone \cite{zhang2024}.

Moreover, the fusion of different data modalities allows for more nuanced and context-aware interactions. Agentic AI can adapt its communication style based on the user's emotional state, cognitive abilities, and cultural background, making the technology more inclusive and effective \cite{adaline2025}. 

\begin{table*}[ht!]
\centering
\caption{Future Research Directions and Priorities for Agentic AI in Elderly Care.}
\label{tab:future_directions}
\begin{tabular}{|p{4.25cm}|p{9.5cm}|p{3cm}|}
\hline
\hline
\textbf{Research Direction} & \textbf{Focus Areas} & \textbf{References} \\ 
\hline 
\hline
Development of Standardized Guidelines and Prompt Design 
& Creation of reproducible and transparent prompt engineering standards; validation protocols for elderly-specific scenarios; training programs for caregivers 
& \cite{perez2025,healthtech2025,ma2023,Asgari2025,clinician2024} \\ \hline
Integration of Multi-modal Information and Features 
& Use of voice, visual, and sensor data for improved interactions; real-time monitoring from wearables and IoT devices 
& \cite{jin2022,mecs2016,wearable2024,zhang2024,adaline2025} \\ 
\hline
Addressing Ethical Concerns 
& Development of AI ethics frameworks; safeguards for privacy, fairness, consent, and algorithm transparency
& \cite{he2024,pmc2024,fhg2024,abadir2024,ageism2024} \\ 
\hline
Adversarially Robust Agentic AI 
& Defense against prompt injection, jail-breaking, and misinformation; secure model architectures and explainability
& \cite{yang2024,lin2025,ferri2025,lin2024large} \\ 
\hline
Proactive and Personalized Support 
& Predictive interventions; learning from user behavior to tailor reminders, care plans, and engagement strategies 
& \cite{ferri2025,zhang2024,adaline2025,mecs2016} \\ 
\hline
Standard Evaluation Frameworks 
& Development of benchmarking protocols; inclusion of user feedback and longitudinal outcomes tracking 
& \cite{ma2023,Asgari2025,clinician2024,abadir2024} \\ 
\hline 
\hline
\end{tabular}
\end{table*}

\subsection{Addressing ethical concerns}
As Agentic AI becomes more autonomous and influential in elderly care, the need to balance innovation with ethical considerations grows increasingly urgent \cite{he2024}. The rapid pace of technological advancement must be matched by robust ethical frameworks that prioritize the rights, dignity, and well-being of older adults. 
For example, Agentic AI often processes sensitive health information, raising concerns about data security and the potential misuse or unauthorized access \cite{pmc2024}.

To address these challenges, research should focus on developing and refining ethical guidelines for the deployment of Agentic AI in elderly care. These guidelines emphasize the importance of informed consent, data minimization, and the right to explanation \cite{fhg2024}. Additionally, mechanisms for human oversight and intervention should be built into the design of these systems, ensuring that critical decisions are always subject to review by qualified professionals. Transparent logging and audit trails can help maintain accountability and build trust among users and stakeholders \cite{abadir2024}.

Another critical aspect is mitigating algorithmic bias in Agentic AI-based solutions for elderly care. Agentic AI is only as fair and inclusive as the data on which they are trained. Biases in training data can lead to unequal care recommendations or the perpetuation of existing health disparities \cite{ageism2024}. Future work should prioritize the development of diverse and representative datasets, as well as tools for ongoing detection and mitigation of bias. By embedding ethical principles into every stage of development and deployment, Agentic AI can help ensure that the benefits of AI-driven care are equitably distributed across all segments of the ageing population.

\subsection{Adversarially Robust Agentic AI}
The increasing autonomy of Agentic AI in elderly care makes them attractive targets for adversarial attacks, where malicious actors attempt to manipulate or deceive the system for harmful purposes \cite{yang2024}. Especially the susceptibility of LLMs to
injection and jail-breaking attacks is a big concern in sensitive applications like healthcare. Ensuring the robustness of these models against such attacks is a critical research priority. Adversarial robustness involves designing LLMs that can withstand attempts to inject misleading information, exploit vulnerabilities in prompt engineering, or bypass security safeguards \cite{lin2025}.

Research in this area should focus on developing advanced techniques for detecting and mitigating adversarial inputs. For example, multi-agent consensus mechanisms can be used to cross-verify outputs and identify inconsistencies or anomalies \cite{ferri2025}. Additionally, continuous monitoring and fine-tuning of models can help maintain their resilience in the face of evolving threats. The integration of explainable AI techniques can further enhance robustness by making it easier to understand and audit the decision-making processes of Agentic AI \cite{lin2024large}.

Another critical aspect of adversarial robustness is protecting sensitive data. Agentic AI must be designed to prevent unauthorized access or leakage of personal health information. Techniques such as federated learning, differential privacy, and secure multi-party computation can help safeguard data while preserving the utility of the models \cite{lin2025}.  
By prioritizing adversarial robustness, the elderly care community can build trust in Agentic AI and ensure their safe and responsible use.

\subsection{Proactive and Personalized Support}
One of the most promising aspects of Agentic AI in elderly care is the move toward proactive and personalized support. Traditional care systems often react to problems as they arise, but Agentic AI has the potential to anticipate needs and intervene before issues become critical \cite{ferri2025}. For example, by analyzing patterns in health data, social interactions, and daily activities, these systems can identify early warning signs of cognitive decline, social isolation, or physical health deterioration \cite{zhang2024}.

Personalization is key to the effectiveness of Agentic AI. Each elderly individual might have unique needs, preferences, and challenges, and care solutions must be tailored to meet these needs accordingly \cite{adaline2025}. Future research should focus on developing adaptive algorithms that can learn from ongoing interactions and adjust their support strategies in real-time. This includes personalized reminders, cognitive exercises, and social engagement activities tailored to the individual's abilities and interests.

Moreover, proactive support can extend beyond health monitoring to include assistance with daily living tasks, such as medication management, scheduling appointments, and managing home automation. Agentic AI can integrate with smart home technologies to automate routine tasks, provide step-by-step guidance for complex activities, and even facilitate social connections through virtual companions  \cite{mecs2016}. The ultimate goal is to empower older adults to maintain their independence and quality of life for as long as possible while providing peace of mind to caregivers and family members.

\subsection{Standard Evaluation Frameworks}
The successful integration of Agentic AI in elderly care depends on establishing standard evaluation frameworks that can assess their performance, safety, and impact \cite{ma2023}. Current evaluation methods are often short-term or fragmented and inconsistent, making it difficult to compare different systems or accurately measure their real-world effectiveness. Standard frameworks should define clear metrics for accuracy, reliability, user satisfaction, and clinical outcomes, as well as procedures for ongoing monitoring and improvement \cite{Asgari2025}.

Moreover, longitudinal studies are required to assess the effectiveness and identify the consequences of Agentic AI in elderly care. Such studies can track outcomes across domains like cognitive health, emotional well-being, societal and economic impact, and last but not least, capture metrics related to trust and digital inclusion. 

Evaluation frameworks should also incorporate mechanisms for user feedback and participatory design, ensuring that the voices of older adults and caregivers are central to the evaluation process. For example, regular surveys, focus groups, and usability testing can help identify areas for improvement and ensure that the technology remains aligned with user needs \cite{clinician2024}. Additionally, evaluation frameworks should include provisions for auditing and accountability, such as transparent logging and independent review.

Ultimately, standard evaluation frameworks must be adaptable enough to keep pace with the rapid advancements in technological innovation \cite{abadir2024}. As Agentic AI evolves and new applications emerge, evaluation criteria and methodologies should be updated accordingly. By establishing robust and adaptable frameworks, the elderly care community can ensure that Agentic AI delivers sustained value and continues to meet the complex and evolving needs of the ageing population.

\section{Conclusions}
Agentic AI represents a significant leap forward in the evolution of elderly care, offering the ability to deliver personalized, adaptive, and proactive support to older adults. By combining autonomous reasoning, contextual understanding, and multi-agent collaboration, these systems have the potential to address the complex and dynamic needs of ageing populations, achieving improvement and scalability. The integration of Agentic AI can enhance care coordination, empower caregivers, and foster greater independence and well-being among seniors. Yet, the transformative promise of Agentic AI will only be realized through a thoughtful and collaborative approach. Researchers, developers, healthcare professionals, caregivers, and policymakers must collaborate to ensure that these technologies are developed and deployed ethically, inclusively, and with profound respect for privacy and human dignity. Addressing challenges, such as data security, algorithmic fairness, and accessibility, must remain at the forefront of innovation. The future of elderly care lies in harnessing the strengths of Agentic AI while remaining vigilant to its risks and limitations. By prioritizing human-centered design, transparent evaluation, and ongoing collaboration across disciplines, we can build a future where advanced AI not only augments healthcare delivery but also enriches the lives of older adults, making care more compassionate, equitable, and responsive for generations to come.

\bibliographystyle{IEEEtran}
\bibliography{Manuscript}

\end{document}